\documentclass[journal]{IEEEtai}

\usepackage[colorlinks,urlcolor=blue,linkcolor=blue,citecolor=blue]{hyperref}

\usepackage{color,array}

\usepackage{blindtext, graphicx}
\usepackage{multirow}
\usepackage{cite}
\usepackage[ruled,vlined,commentsnumbered,linesnumbered,lined,boxed]{algorithm2e}
\usepackage{algorithmic}
\usepackage{amsmath}
\usepackage{xcolor}
\usepackage{soul}
\usepackage[bb=px]{mathalfa}

\newcommand{\hlr}[1]{{\textcolor{black}{#1}}}

\begin{document}

\title{Reduction of Class Activation Uncertainty with Background Information}

\author{H M Dipu Kabir.
\thanks{H M Dipu Kabir is with Charles Sturt University, Australia. (Email: hmdkabir@connect.ust.hk)}

\thanks{This paragraph will include the Associate Editor who handled your paper.}}

\markboth{IEEE Transactions on 
}
{First A. Author \MakeLowercase{\textit{et al.}}: Bare Demo of IEEEtai.cls for IEEE Journals of IEEE Transactions on Artificial Intelligence}

\maketitle

\begin{abstract}
Multitask learning is a popular approach to training high-performing neural networks with improved generalization. In this paper, we propose a background class to achieve improved generalization at a lower computation compared to multitask learning to help researchers and organizations with limited computation power. We also present a methodology for selecting background images and discuss potential future improvements. We apply our approach to several datasets and achieve improved generalization with much lower computation. Through the class activation mappings (CAMs) of the trained models, we observed the tendency towards looking at a bigger picture with the proposed model training methodology. Applying the vision transformer with the proposed background class, we receive state-of-the-art (SOTA) performance on \hlr{CIFAR-10C,} Caltech-101, and CINIC-10 datasets. Example scripts are available in the `CAM' folder of the following GitHub Repository: github.com/dipuk0506/UQ
\end{abstract}

\begin{IEEEImpStatement}
A well-generalized machine-learning model provides good performance in both training and test datasets. In this paper, we have proposed an efficient generalization technique. The proposed method generalizes both the initial layers and the end layers. Moreover, the proposed method requires a smaller model and less training time compared to multitask learning. We applied the proposed method with transfer learning, vision transformers, and SpinalNet end layers and achieved eye-catching performance. We achieved state-of-the-art performance in \hlr{CIFAR-10C,} Caltech-101, and CINIC-10 datasets. We shared scripts and made the paper publicly available. Researchers have started to apply our method and cite our paper. We can potentially observe more applications of the proposed method in the upcoming years.
\end{IEEEImpStatement}

\begin{IEEEkeywords}
Vision Transformer, \hlr{Large Language Model}, Ablation Study, Transferred Initialization, Class Activation Map, Background Class.
\end{IEEEkeywords}

\section{Introduction}

\IEEEPARstart{A}{lthough}  deep machine learning models have brought revolutionary performance in computer vision, deep learning models often perform poorly on test data due to poor generalization \cite{caruana2000overfitting}. The classification score for each class is a weighted sum of the final convolutional layer outputs. Researchers have recently observed that some spatial positions (x,y) of the last convolutional layer contribute a high value toward the score for an image \cite{zhou2016learning} \hlr{for a class.} Such a spatial position (x,y) usually contains a pattern of that class. That brings us the opportunity to see the effect of performance enhancement techniques on recognizing individual features.

\begin{figure}
\begin{center}
\centerline{\includegraphics[clip, trim=2.2cm 0.9cm 1.6cm 0.5cm, width=\columnwidth,angle=0]{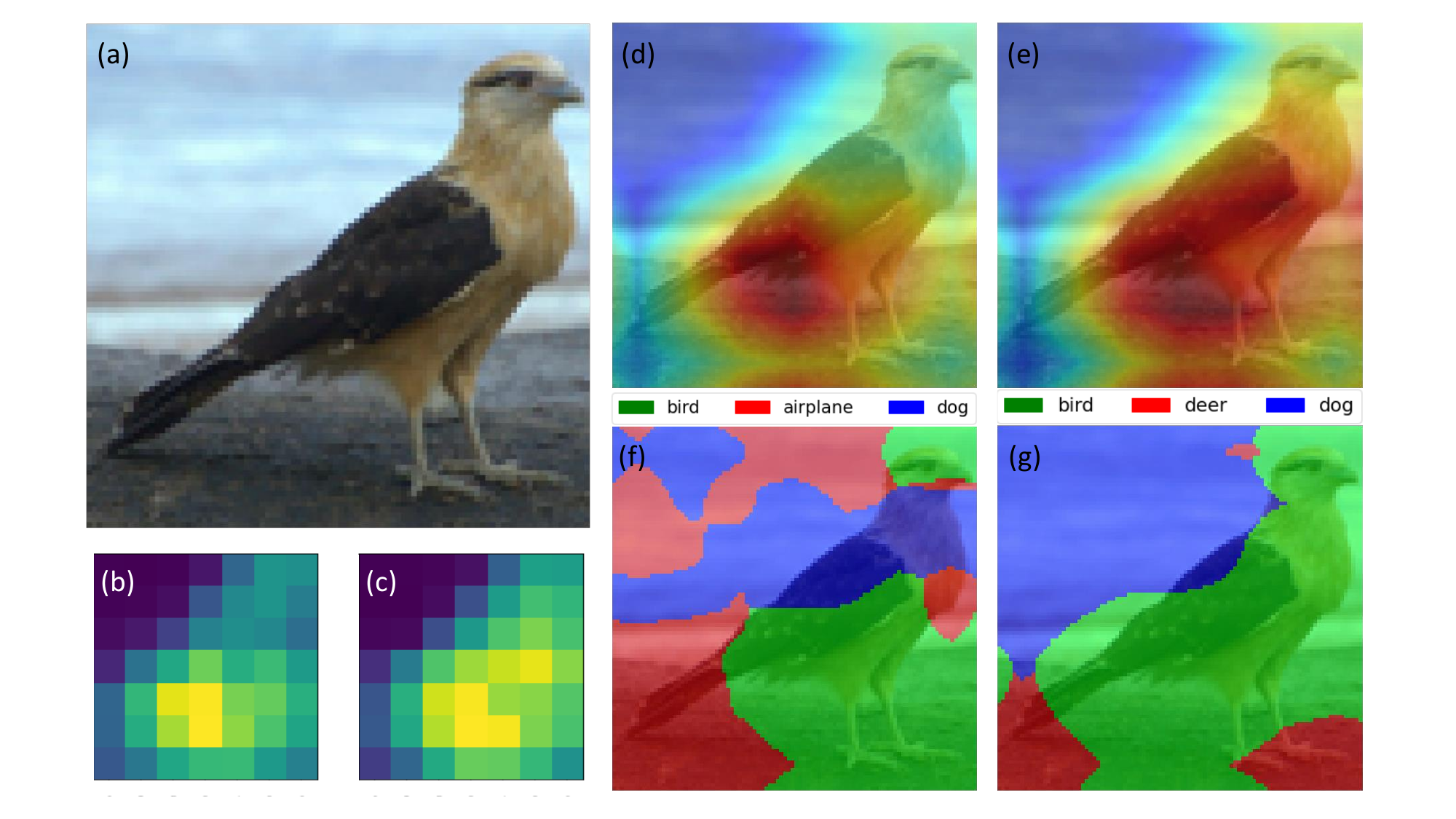}}
\caption{Models with both traditional training and training with background class are applied to a bird image (a) in the STL-10 dataset. Subplots (b) and (c) show class activation mapping of the bird class on the final convolutional layer respectively for traditional and for training with background situations. Subplots (d) and (e) show class activation mapping with the image. Subplots (f) and (g) show deep feature factorization results on the image for the traditional and proposed method respectively.}
\label{Bird}
\end{center}
\end{figure}

Uncertainties in the deep learning model originate from deep layers and how the final head layer interprets the information from the last convolutional layers. The performance of transfer-learned models also varies depending on the structure of the models \cite{kolesnikov2020big}. Therefore, simpler deep neural networks (DNNs) may not propagate important features of a few classes in a large dataset \cite{zagoruyko2016wide}. However, even larger DNNs or DNNs of moderate size do not ensure good performance. One common reason for their poor performance is prioritizing less relevant features instead of important and robust patterns. Fig. \ref{Bird} shows one such situation and the effect of our proposed improvement. Fig. \ref{Bird}(a) shows the image of the bird. Fig. \ref{Bird}(b) \hlr{and Fig. \ref{Bird}(c)} shows class activation maps (CAMs) respectively for the traditional and the proposed model. Later subplots present CAMs on images and deep feature factorization on the image. A bird might be sitting and its eyes can be closed in a test image. A \hlr{neural} network deciding the class of an image by seeing most portions of a bird is expected to be more robust than a \hlr{neural} network, predicting by seeing the legs and eyes on an image \cite{kapidis2019multitask}.

Transfer learning and multitask learning are two revolutionary approaches toward improved generalization \cite{kolesnikov2020big, gesmundo2022evolutionary}. Transfer learning is a powerful technique for researchers and institutions with low computation resources. Although initial layers are generalized in transfer learning, randomly initialized head layers with inadequate training can cause a high generalization error. Multitask learning can potentially generalize both initial and final layers and has brought superior performance in several datasets \cite{gesmundo2022evolutionary}. Transfer learning with the help of publicly available pre-trained models requires much lower computation. Multitask learning requires much more computing compared to transfer learning and learning from random initialization. Moreover, multitask learning requires more attention to the data. While training NN for both handwritten digits and handwritten characters of different languages, images of two different classes can be very similar. Such as the number `0', and the letter `O' are similarly written by many participants. Moreover, the accuracy of the model in multitask learning can potentially degrade while training a shared layer for two completely different tasks \cite{kokiopoulou2021flexible, standley2020tasks}.

Both the initial layers and head layers need good generalization to achieve good performance. Several models with highly generalized initial layers are publicly provided by reputed organizations. Therefore, we try to bring the benefit of both transfer learning and multitask learning with optimal computation. We develop a background class to bring better generalization in head layers.

Large language models (LLMs) have brought several recent SOTA performances in several fields. Researchers are applying LLMs to new and new fields and getting SOTA performances. LLMs are also known as transformers. Dosovitskiy et al. presented a computer vision version of the transformer and received several SOTA \cite{dosovitskiy2020image}. That computer vision version of the transformer is known as the vision transformer. We have also investigated our approach with transformers and received SOTA or near SOTA performances.

\section{Background}
We validate the advantage of introducing a background class with both theory and practical results.  The background class brings improved generalization. Several other approaches can also bring improvement in generalization. As the proposed method brings a similar effect to multiclass learning, we present multitask learning in this section. Class activation mapping (CAM) is an effective tool to evaluate potential reasons for any poor performance. \hlr{This section also presents the use of the background class in existing research areas.}

\subsection{Generalization}

Generalization error is also known as the out-of-sample error or the overfitting error in statistical learning. Both the training and test datasets have a finite number of samples. Moreover, the test samples are different from the training sample, unless there is an overlap of data. An overfitted model on the training data can potentially provide a significantly low accuracy on the test dataset \cite{bousquet2002stability}. One of the popular estimates of generalization error is the \emph{leave one out error}, and another one is the \emph{empirical error}. 


Researchers compute the \emph{leave one out error} by removing one training sample from the training set. They test the trained neural network on the removed sample. Although the \emph{leave one out error} process is computationally extensive, researchers regard it as the most unbiased estimate of the generalization error \cite{chapelle2002choosing}.

Devroye et al.   \cite{devroye2013probabilistic} presented the empirical error with the example of classification. The probability of misclassification becomes the error $l(f(),z_i) = P(f(x_i) \neq y_i)$, where, $P()$ is the probability function, $f()$ is the model function, $z_i = (x_i,y_i)$ is the input-output combination. More details on empirical risk management in classification problems are discussed in section 4 of \cite{devroye2013probabilistic}.


Both of the error estimates are derived from $\mathbb{E}[l(f(),z)]$. Where,  $\mathbb{E} [.]$ is the expectation function, $l()$ is the loss function, $f()$ is the model function and $z$ is the sample \cite{bousquet2002stability}. $\mathbb{E}[l(f(),z)]$ is known as the \emph{generalization error} and expressed as:
\begin{equation}
\mathbb{E}[l(f(),z)] = \frac{1}{m} \sum_{i=1}^m l(f(),z_i),
        \label{GE_Eq1}
\end{equation}
where $m$ is the number of samples in the dataset. Regularization is a popular technique for generalization improvement \cite{micchelli2005learning}. The cost function for regularized model training has the following form: 
\begin{equation}
\underset{f()}{\text{min}} \ \ \frac{1}{m} \sum_{i=1}^m l(f(),z_i) + \lambda \sum_{j=1}^p |w_j|,
        \label{GE_Eq2}
\end{equation}
where $w_j$ is the $j^{th}$ parameters of the model $f()$ and $p$ is the number of parameters in $f()$. The research community considers high $|w_j|$ values as an indication of sharp changes in predictions over the input domain, resulting in overfitting. Therefore,  high $|w_j|$ values are penalized in the cost function.

Several other common approaches to reducing generalization error are: using more training data \cite{caruana2000overfitting}, data augmentation \cite{cubuk2019autoaugment}, early stopping \cite{caruana2000overfitting}, optimal model structure \cite{hernandez2023training}, neural architecture search \cite{pham2018efficient}, dropout \cite{gal2016dropout}, cross-validation \cite{efron1983estimating, krogh1994neural}, pre-trained model \cite{kolesnikov2020big}, and multitask learning \cite{ndirango2019generalization}. Besides these common approaches, many researchers have proposed novel approaches to improve generalization. 

\subsection{Multitask learning}
Multitask learning can potentially improve generalization and bring superior performance when tasks are related. Several researchers have provided theories indicating improved generation in multitask learning. According to Ghifary et al., \cite{ghifary2015domain} multitask learning autoencoder training tries to minimize both \hlr{shared} weights ($|w|$) and individual task-specific weights ($|v^t|$), where $t$ is the task number ($t=1,2,3... N_T$). As they regularized $|w|$ for all tasks in a loop, the trained model becomes more generalized.

\begin{figure*}[ht]

\begin{center}
\centerline{\includegraphics[clip, trim=0.3cm 7.4cm 3.5cm 1.8cm, width=6.5in,angle=0]{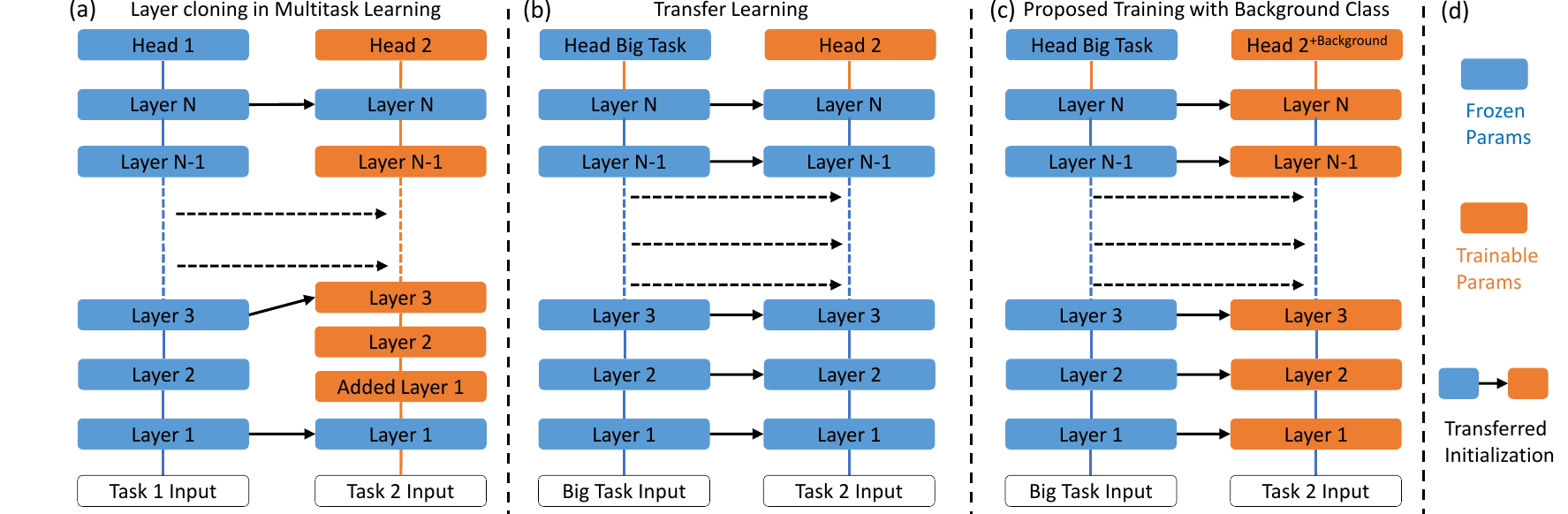}}
\caption{Rough diagrams presenting multitask learning, transfer learning, and proposed training with one or more background classes. In multitask learning (a), the child model is created from a parent model. Layers can be frozen, and the initialization of non-frozen layers can be transferred or randomly initialized. New layers can be added or removed. \hlr{In traditional transfer learning (b), all initial layers are usually kept frozen.}  In proposed work (c), we start with a transfer learned model. \hlr{Initial parameter values are copied from the pre-trained model. Layers are not frozen during the training.} The head contains classes of the data and the background class.  }
\label{Chart}
\end{center}

\end{figure*}

Multitask learning can statistically keep good activation values over a larger space (x,y) in the final convolutional layer \cite{kapidis2019multitask}. Such DNNs often recognize the class by seeing a larger portion of the image. When a model recognizes a bird by seeing only legs, the model may misclassify an image where a bird is sitting. Moreover, some tree branches may appear quite similar to birds' legs. Therefore, the prediction by seeing a larger portion of the bird is more robust than the prediction by seeing only the leg.
According to Ndirango et al., \cite{ndirango2019generalization} multitask learning can potentially bring improved generalization through the conditional improvement in loss function values.
Fig. \ref{Chart}(a) presents several possible combinations of multitask learning. We perform multitask learning without freezing any layer. The network receives pre-trained weights from a previously trained model. All the parameters are changeable during the training. We consider only classes of the investigated dataset while testing. Several researchers have recently proposed improvements and new concepts in multitask learning \cite{lian2022scaling, li2023provable, ghiasi2021multi}. \hlr{Researchers can potentially apply those approaches in the future to achieve superior performance and to perform more detailed analyses.}

\begin{figure}[ht]

\begin{center}
\centerline{\includegraphics[clip, trim=0.3cm 10.2cm 20.9cm 1.8cm, width=\columnwidth,angle=0]{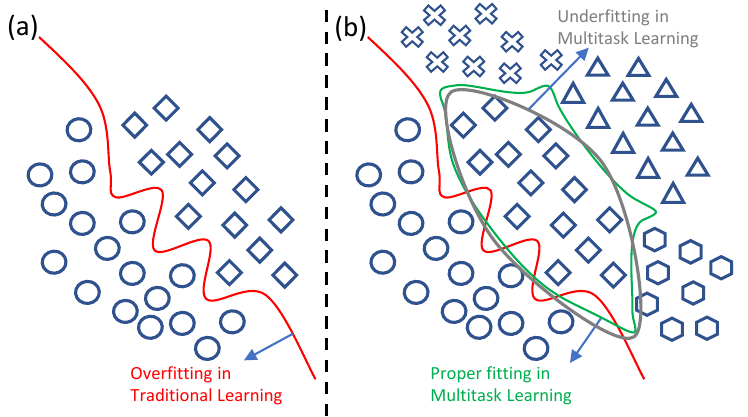}}
\caption{Rough diagrams to explain potential generalization improvement in multitask learning.  The decision boundary in traditional learning (a) becomes overfitted. The decision boundary in multitask learning can potentially be more generalized (b) than the traditional one. However, multitask learning can potentially bring underfitting issues.}
\label{Cluster}
\end{center}

\end{figure}

Fig. \ref{Cluster} presents two rough diagram visualizations to explain potential generalization improvement in multitask learning. Deep learning models have a large number of parameters. A few training samples do not generalize parameters. The decision boundary in traditional learning, shown in Fig. \ref{Cluster} (a) becomes overfitted. A slight variation in test data can potentially result in very low accuracy. The decision boundary in multitask learning can potentially be more generalized than the traditional one. In Fig. \ref{Cluster}(b), a proper decision boundary is depicted by a green contour. The Grey contour represents a decision boundary where the model is facing an underfit issue.

Multitask learning can bring overfitting, underfitting, or proper fitting based on the added dataset. Overfitting occurs when the dataset is small compared to the model complexity. Adding another classification dataset \hlr{up to a limit} increases generalization. However, if we add many datasets of different types, the model fails to express many complex relationships of the target classification problem \cite{ng2020coursera}.  
\hlr{The amount of data for each task in multitask learning needs to be quite similar for successful multitask learning} \cite{ng2020coursera}. The proposed background class also brings improvement in generalization through data injection in a new class. Proper balancing of data size is needed for good performance.

\subsection{Class Activation Mapping}
Researchers have recently observed that convolutional neural networks can detect the class activity regions in the image besides providing an overall classification result \cite{zhou2016learning}. The spatial $(x,y)$ values of the last convolutional layer get multiplied by the weights of the fully connected layer and added toward the score of each class ($S_c$). The following equation presents the simplest form of computing score from the last convolutional layer: 
\begin{equation}
S_c = \sum_{x,y} M_c (x,y),
        \label{CAM1}
\end{equation}
where, $M_c(x,y)$ is the class activation map value for $(x,y)$ location at the last convolutional layer. $M_c(x,y)$ is computed as follows:
\begin{equation}
M_c (x,y) = \sum_{k} w_k^c f_k (x,y),
        \label{CAM2}
\end{equation}
where, $f_k (x,y)$ is the activation unit $k$ at $(x,y)$ location at the last convolutional layer.

\subsection{Ablation Study}
\hlr{Ablation studies are widespread methods to understand the influence of individual components of neural networks.} The concept of ablation study came from biology. Traditional biological ablation studies include removing components from an organism and investigation. Researchers in the machine learning community are also performing ablation studies to investigate the contribution of individual components \cite{sheikholeslami2019ablation}. 

Several recent research papers presented a relationship between the result of ablation and good generalization \cite{morcos2018importance, ramaswamy2020ablation, zhou2018revisiting}. Networks containing weights of high magnitudes ($|w_j|$) are usually poorly generalized. The removal of a single component ($j$) can potentially cause a high variation in results in such a poorly generalized network. Dropout, batch normalization, and other generalization techniques make models robust against ablations through an improved generalization \cite{morcos2018importance}.

\begin{figure*}[ht]

\begin{center}
\centerline{\includegraphics[clip, trim=0.3cm 3.2cm 2.9cm 1.6cm, width=5in,angle=0]{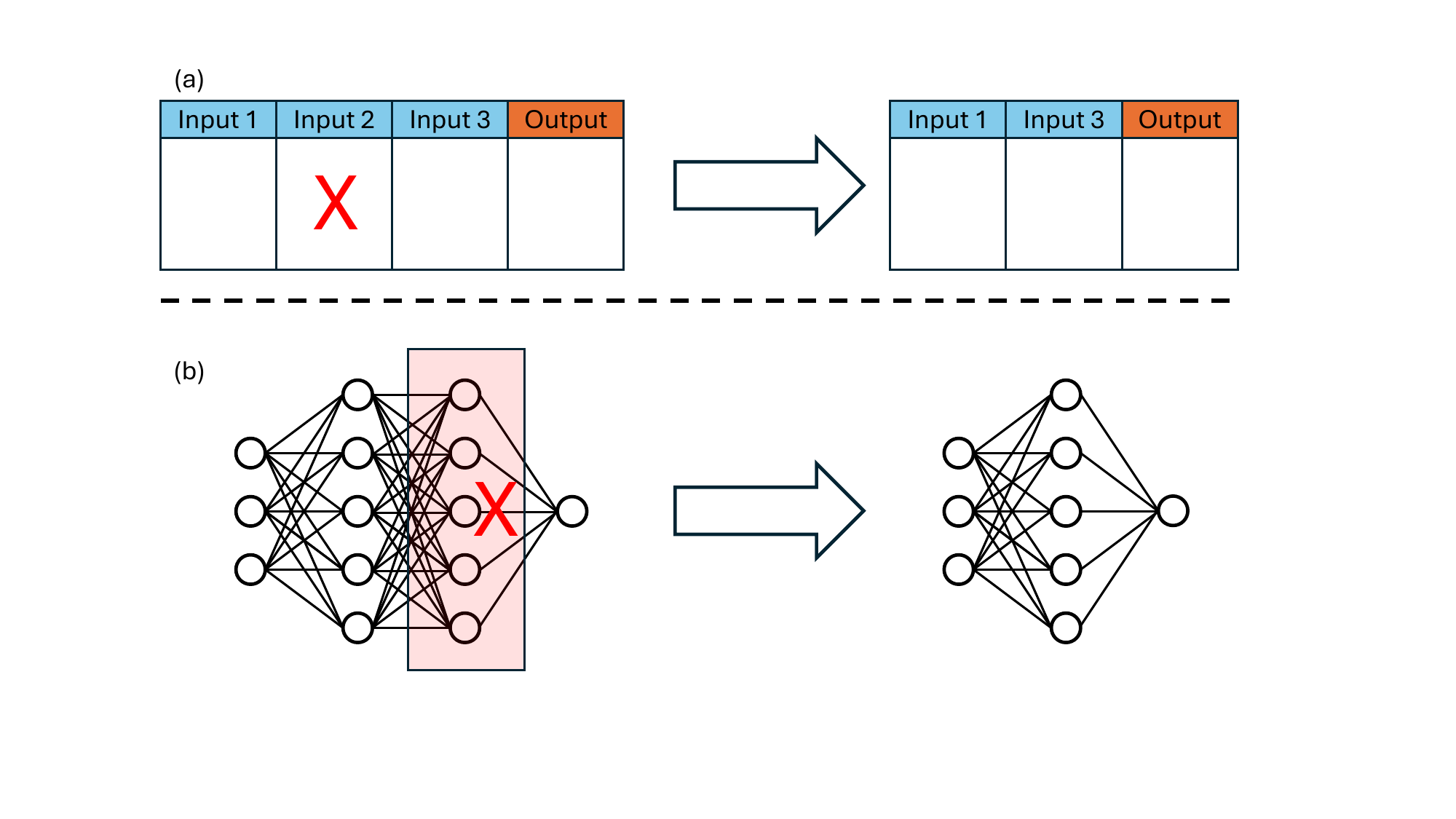}}
\caption{Visualization of the Ablation Study. (a) Data Ablation: One input parameter is removed to observe the effect. (b) Model Ablation: A portion of the model is removed to observe the effect.} 
\label{Ablation}
\end{center}

\end{figure*}

Fig. \ref{Ablation} presents a rough visualization of ablation studies. Researchers practice two types of ablations in the machine learning domain: 1) Data Ablation and 2) Model Ablation. Data Ablation is also known as Feature Ablation.  Fig. \ref{Ablation}(a) presents an example Data Ablation technique. Researchers remove one or more input parameters and observe the effect in a Data Ablation experiment. Researchers remove one or more model components to observe the effect in a Model Ablation experiment. Fig. \ref{Ablation}(b) presents an example Model Ablation technique. A well-generalized model or training methodology faces comparatively low degradation in performance due to the ablation. We draw the figure according to the concept of ablation study presented in \cite{sheikholeslami2019ablation}. 
\hlr{ Researchers have investigated the ablation of all layers. We apply pre-trained models. Any change in the pre-trained weights in the initial layer restricts the propagation of important features to later layers. Therefore, we ablated only the last convolutional layer to observe the difference. }

\subsection{Previous Use of Background Class}
The use of background class is common in detection-type problems. In such problems, the algorithm aims to detect one pattern and other patterns are considered as the background. The concept of background class is previously applied in concrete detection \cite{son2012automated}, defect detection \cite{park2016machine}, ribonucleic acid (RNA) detection \cite{bacstanlar2014introduction} and energy physics \cite{whiteson2009machine}. Son et al. \cite{son2012automated} collected 108 images of concrete surfaces and other surfaces. They transformed red-green-blue (RGB) images to non-RGB color spaces and applied several models. Park et al. \cite{park2016machine} proposed an approach for the automatic detection of wear, burrs, scratches, and dirties. They train CNNs for detection. They claimed time, cost-effectiveness, and higher accuracy than humans. According to Bacstanlar et al., \cite{bacstanlar2014introduction}, the RNA detection problem can also be similar to the detection problem where the desired pattern belongs to the target class and all other patterns belong to the background class.
Whiteson et al. \cite{whiteson2009machine} applied supervised machine learning with background classes for high-energy physics. They collected data from the Fermilab Tevatron accelerator and developed a dataset with five background classes. They applied a fully connected shallow neural network and received state-of-the-art (SOTA) results at the time when the work was published. 

The Caltech-101 dataset has a background class \cite{li2004caltech}, but the background class of that dataset contains a few distinctive images. Moreover, several background images of that dataset contain human faces that overlap with another class. Therefore, we develop background classes by ensuring that there is no common pattern.

\subsection{\hlr{Domain Adaptation and Domain Generalization}}
\hlr{Human brains can see an object as logos, posters, cartoons, sketches, etc., and detect the object from natural images. The Eiffel Tower is a popular iconic building. We often see sketches, logos, and paintings of the Eiffel Tower. Our brain can easily detect the image of the Eiffel Tower from different sources. Researchers are improving NNs for detecting objects in new domains \cite{peng2019moment}. Such robustly trained neural networks are useful in detecting objects in adverse situations. For example, autonomous driving requires help from NNs. A traditionally trained NN may not work well in adverse weather. The weather can be rainy, cloudy, or snowy. People may prefer to travel at night when most training data contain images during the daytime. A domain-adaptable NN can perform well in unknown situations \cite{wang2022continual, chen2022contrastive}.}

\begin{figure}[ht]

\begin{center}
\centerline{\includegraphics[clip, trim=3.6cm 12.0cm 25.7cm 0.2cm, width=6.6cm,angle=0]{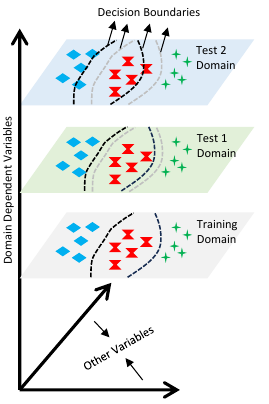}}
\caption{\hlr{Shift of decision boundaries due to change in domains. A poorly trained NN can potentially perform well in the training domain and may perform poorly in test domains.}}
\label{Adaptation}
\end{center}
\end{figure}

\hlr{Fig. \ref{Adaptation} presents how trained NN can perform poorly in a new domain. The vertical axis of this diagram presents all variables related to domain shifts and not related to class objects. Two horizontal axes present all variables not related to the change in domain. The situation assumes that there is no variable related to both domain and classification objects. Target objects are grouped into three classes. A trained NN performs well in the training domain. Black dotted lines in the training domain present decision boundaries. Grey dotted lines in other domains present the positions of actual decision boundaries in those domains. The decision boundary changes over the domain due to improper training. Black dotted lines in test domains present shifted decision boundaries in those domains. That shift can potentially cause misclassification for many objects. We are showing only the shift of decision boundary due to changes in domains. The boundary also may get rotated or follow other transformations due to changes in domains.}

\hlr{Domain Generalization and Domain Adaptations are two closely related research areas \cite{wang2022generalizing}. The target domain is unknown in domain generalization, where the target domain information is available in the domain adaptation. Domain generalization is more challenging and more effective for practical applications compared to domain adaptation. Multitask learning, image augmentations, transfer learning, etc. methods improve domain generalization \cite{zhou2022domain}. The proposed method can potentially improve domain generalization. }

\section{Contribution}
We start this section with our observations and hypotheses. After that, we propose the background class to \hlr{improve generalization} in head layers. As we \hlr{consider} initial layers to be transfer learned, they contain well-generalized parameters. Therefore, training head layers with more samples can potentially improve the generalization of head layers. \hlr{According to the theoretical explanations and analysis in this section, the background class can potentially get a better generalization and restrict irrelevant background patterns} from providing a high score to a class. 

\subsection{Observations}
We reached several conclusions based on our observations. These conclusions helped us to find the idea of adding a background class. Other groups of researchers may agree or may disagree with our conclusions. Therefore, we write our conclusions as hypotheses. Observations with hypotheses are as follows:  

\hlr{Observation 1: Initial layers of models pre-trained on a similar large dataset can propagate almost all essential features of the target dataset without further training. Post-trained head computes score-portions $M_c (x,y)$ based on recognized features for different classes $(C)$.} Yosinski et al. \cite{yosinski2014transferable} performed a study on the transferability of features. They achieved similar findings. Later, Zhou et al. \cite{zhou2016learning} observed how the final layer computes scores from $M_c (x,y)$.

\emph{Hypothesis 1: Pre-trained models on a large dataset can be applied to classify new images of \hlr{similar or} slightly different types with a post-trained head, and such combinations often bring eye-catching performance. However, when new images are quite different from the dataset of pre-training, the training can bring a poor performance.} 

\hlr{Observation 2:} Although DNNs have a large number of parameters, they can potentially face underfitting issues in very large datasets. Torchvision \cite{marcel2010torchvision} provided several pre-trained models with the ImageNet accuracy information. ResNet18 models have received much lower accuracy on the ImageNet dataset compared to the ResNet50 model. Different model structures with the same parameter count usually have different accuracy on the same dataset. However, the accuracy of the two models can be the same, accidentally. The propagation of features ($f_k (x,y)$) and decoding may depend on the structure of the model. However, artificial intelligence is still a growing field. Researchers may find a better training method where ResNet18 can potentially provide comparable accuracy with bigger models \cite{wightman2021resnet}. 

\emph{Hypothesis 2: Different DNN models usually exhibit different accuracies for the same dataset and \hlr{the same} training conditions. \hlr{The propagation of features ($f_k (x,y)$) and decoding can potentially depend on the structure of the model.} }

\hlr{Observation 3:} Fully connected head layers can provide a high $M_c (x,y)$ to one class by decoding several features. \hlr{The layer can also provide a high $M_c (x,y)$ value by detecting the eye and legs of a bird at two different $(x,y)$ locations. We often observe} high $M_c (x,y)$ values when positions $(x,y)$ of features are interchanged. In Fig. \ref{Bird}(f), the eye and legs provide the highest score to the bird class. \hlr{The fully connected layer can potentially decode some features incorrectly.} The white background provides the highest score for the airplane class. The dorsal area of the bird indicates the dog class. Different features are propagated parallelly over the 2D convolutional planes until the last convolutional layer. Each convolutional layer also performs a modulation based on weights and activation functions. 

\emph{Hypothesis 3: Convolutional layers propagate different features with modulations parallelly in spatial space (x,y). The fully-connected head of the DNN model computes scores for classes based on the modulated output features of the last convolutional layer.}

\hlr{Observation 4:} While detecting a bear in terrain, the terrain is the background. However, the model can potentially detect that terrain as a rugby ball\footnote{https://jacobgil.github.io/pytorch-gradcam-book}. That can potentially happen due to the presence of terrain in many images in the training set, containing a rugby ball. In Fig. \ref{Bird}(f), the white background provides a high score for airplane class. That can potentially happen due to the presence of a white background in many images in the airplane class.

\emph{Hypothesis 4: Features of background objects also contribute to scores of different classes. When a significant number of training images of one class contains certain background objects, the head of the model decodes the features of those background \hlr{objects as features} of that class.}

\hlr{Observation 5: In traditional transfer learning, all the initial layers are frozen.} Therefore, $f_k (x,y)$ becomes the same when the image and image augmentation are the same. The head can be trained \hlr{on} different datasets and with a different set of training parameters. However, those training do not change $f_k (x,y)$. However, researchers have also investigated transfer learning without freezing all initial layers and by unfreezing initial layers after certain epochs \cite{goodfellow2016deep, guo2019spottune}.

\emph{Hypothesis 5: When all the initial layers are frozen in the transfer learning, the value of $f_k (x,y)$ is a function of the image, augmentation, and pre-trained weights. Training of the head optimizes parameters ($w_k^c$ and biases) in the head.}

The number of parameters in a single layer fully connected head network $N_P^H$ is directly proportional to the number of class $N_c$ (i.e. $N_P^H \propto N_c$). Weight increases with the increase of both the number of features and the number of classes in a proportional manner. The number of biases on the fully connected head layer is equal to the number of classes. 
The multitask learning method trains models with several classification problems. Therefore, the training consists of training different heads for different datasets. The parameter in each head layer of subnetworks $N_P^H$ is directly proportional to the number of class $N_c$ of the corresponding dataset. The total number of trainable head parameters increases drastically with multitask learning. The number of samples also becomes the summation of individual sample numbers. Both the data and the number of trainable parameters increase the multitask learning by several times compared to traditional transfer learning. The Graphics processing unit (GPU) or Tensor processing unit (TPU) requirement also increases with increased parameters. The batch size of DNN training needs to be reduced to meet the size of the GPU. Therefore, the training time increases significantly.

\subsection{Improvement with Background Class}
Before the invention of class activation mapping, researchers proposed to reduce patch distance \cite{zuo2014learning, frome2006image} to improve classification results. Forme et al. \cite{frome2006image} defined patch distance with the help of two terms: \emph{focal image} ($\mathcal{F}$), and \emph{a candidate image} ($\mathcal{I}$). The focal images are images in the training set, and the candidate image is the test image. They define image-to-image
distance function $\mathcal{D}(\mathcal{F},\mathcal{I})$ as follows:
\begin{equation}
\mathcal{D}(\mathcal{F},\mathcal{I}) = \ <w^\mathcal{F} . \ d^\mathcal{F} (\mathcal{I})>
        \label{PDt}
\end{equation}
where, $w^\mathcal{F}$ is a vector of weights, $d^\mathcal{F} (\mathcal{I})$ is the patch distance vector between $\mathcal{F}$ and $\mathcal{I}$.
They consider two different types of images: dissimilar images ($\mathcal{I}^d$) and similar images ($\mathcal{I}^s$). For accurate prediction $\mathcal{D}(\mathcal{F},\mathcal{I}^d)$ $>$ $\mathcal{D}(\mathcal{F},\mathcal{I}^s)$. Their cost function tries to increase the difference between $\mathcal{D}(\mathcal{F},\mathcal{I}^d)$ and $\mathcal{D}(\mathcal{F},\mathcal{I}^s)$. Their cost function also regularizes weights ($||w^\mathcal{F}||^2$).

Similarly, the feature activation unit $f_k (x,y)$ can potentially represent one class (c). The optimization increases $w_k^c f_k (x,y)$ over iteration when $f_k (x,y) \in \mathcal{F}_k^c (x,y)$. Where, $\mathcal{F}_k^c (x,y)$ is the superset of potential activation unit values for features representing class $c$ near $(x,y)$ location. The optimization also decreases $w_k^c f_k (x,y)$ over iteration when $f_k (x,y) \notin \mathcal{F}_k^c (x,y)$. In a traditional multiclass classification problem, $\mathcal{F}_k^c (x,y)$ contains both relevant activations from objects and irrelevant activations coming from backgrounds. Let, $\mathcal{F}_k^b (x,y)$ be the superset of potential activation unit values for features representing the background. One class may contain a similar background in a good portion of training samples. The optimization process may consider such a feature as a part of that class object. 

In the proposed method,  the weights to background activation unit values $\{f_k (x,y) \in \mathcal{F}_k^b (x,y)\}$ get reduced over iteration. \hlr{Some features may exist in both the background class and in a target class.} Weight to \hlr{such a} common activation unit values $\{f_k (x,y) \in \mathcal{F}_k^b (x,y)\} \cap \{f_k (x,y) \in \mathcal{F}_k^c (x,y)\}$ are also reduced when there exist more example in the background class. \hlr{According to our observation on deep feature factorization,} the presence of an extra background class with an optimal number of background samples also optimizes $||w_k^c||$. Instead of a very high $ w_k^c f_k (x,y)$ in a small region, and low $ w_k^c f_k (x,y)$ \hlr{elsewhere,} $ w_k^c f_k (x,y)$ becomes high over a larger region of the class object in most samples. 

\hlr{NNs may face underfitting or overfitting issues based on the size of the dataset and the NN structure. In the proposed method, we can adjust the background data. Through trial and error with background images of different types and numbers,} we can optimize the performance of NN.

\subsection{Background Class Generation}
\label{BG_gen}
We follow the following principles for the generation of background class:
\begin{itemize}
  \item Images in the background class should not contain any object, which belongs to a class of the classification problem. 
  \item Images in the classification dataset often contain many patterns in the background. It may not be possible to construct a background class containing all of those background patterns. However, the background dataset developer should try to cover common patterns.
 \item The background class may contain a few monochromatic images so that, the learned models do not give a classification result by seeing only the color. 
 \item The background class may contain some textures that are irrelevant to the target classes so that, the learned models do not get overfitted in the texture domain. 
  \item The number of images in the background class should be suitable for the classification task. 
\end{itemize}
The designer of the background class needs to investigate several images of the classification problem for different classes. For example, the FGVC Aircraft dataset contains images of different types of aircraft. The background of images contains sky, ground, buildings, trees, runway, etc. The designer has to find several other datasets containing those background patterns to construct the background class. For each problem, we develop a background class containing distinctive images from publicly available several datasets.
The selection of the size of the background class also needs several considerations. A background class with a few images may not bring a good generalization. Also, a large background class may create problems of highly imbalanced data distribution over classes. According to our investigation, an acceptable range of the background class can be the size of individual classes to the size of the classification dataset. The background class needs to be small for highly uncertain classification problems. When the trained model without a background class receives very low classification accuracy, adding a background class of a large size can potentially bring poor performance. For example, a trained model without a background class may bring about 70\% overall accuracy on a highly uncertain multiclass classification problem. When 90\% images on the dataset are background images after adding the background class, predicting all samples as the background can bring 90\% accuracy. The trained model may predict all samples as the background. \hlr{The classification accuracy among target classes becomes low.}

\subsection{Weights to the background class on the head layer may not capture all background features}
\hlr{The background class contains many images of different features.} The fully-connected layer of the model may not create optimal weight ($w_k^b$) for assigning high values for all background features $\mathcal{F}_k^b (x,y)$. However, the optimization algorithm can limit ($w_k^c f_k(x,y)$) values for common features existing in backgrounds of class images ($\{f_k (x,y) \in \mathcal{F}_k^b (x,y)\} \cap \{f_k (x,y) \in \mathcal{F}_k^c (x,y)\}$). Images of a class usually contain different background features $\{f_k (x,y) \in \mathcal{F}_k^b (x,y)\}$ in different images. As the background class contains a large number of background images, ($w_k^c f_k(x,y)$) values for features $\{f_k (x,y) \in \mathcal{F}_k^b (x,y)\}$ get reduced for the optimization of the loss function over iterations.

The training stage considers the extra background class, as shown in Fig. \ref{Chart}(c). However, we do not consider the score on the background class ($S_b$) in the test phase. As the problem is a classification problem, we consider the index of the maximum $S_c$ value as the predicted class.

\begin{figure*}

\begin{center}
\centerline{\includegraphics[clip, trim=3.7cm 8.1cm 7.0cm 2.6cm, width=5.2in,angle=0]{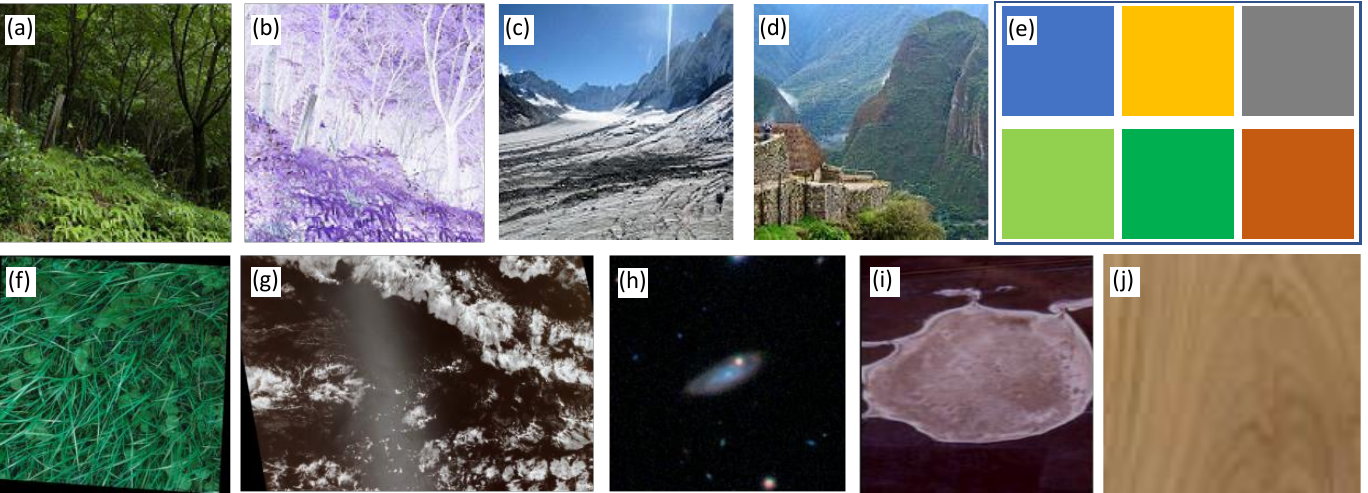}}
\caption{A few example images in the background class for color images: (a) image of a forest, (b) color inverted image of (a), (c) glacier, (d) mountain, (e) monochromatic images, (f) grass, (g) clouds from a satellite image, (h) image of Galaxies, (i) water body from a satellite image, and (j) wood texture. }
\label{Background}
\end{center}

\end{figure*}

\subsection{Class Activation Uncertainty Score}
In this study, we haven't proposed any score for class activation uncertainty. We concentrate on getting larger deep feature factorization regions for the predicted class. Many researchers are trying to develop robust scores for uncertainties in different machine learning domains \cite{marin2016prediction, kendall2018multi}. However, most scores depend on certain assumptions. Therefore, some other groups of researchers could not completely agree with them and developed another score for training and evaluating models \cite{pearce2018high, kabir2021optimal}. Future researchers may find a widely accepted uncertainty score for class activation and uncertainties in other machine learning domains.

\subsection{Training Methodology of Investigated Models}
The multitask learning method can follow complex training combinations \cite{gesmundo2022evolutionary}. The multitask learning model training can freeze several layers and retrain several other layers. It is possible to introduce new layers, omit layers, and train existing layers from random initialization. Fig. \ref{Chart}(a) presents a graphical representation of multitask learning. In our investigated multitask learning, we define a shared fully connected layer containing outputs of different datasets. Previous layers are already well generalized on a large dataset.  Therefore, we declare a common fully connected layer for multiple classification problems for a better generalization of weights on that layer.
The class number on the fully connected layer becomes equal to the summation of class numbers of different datasets. Datasets are also arranged accordingly for the model training. \hlr{Each class of each dataset gets a distinct class identity.}

\hlr{In traditional transfer learning, all initial layers are usually frozen. The head layer is structured according to the output format of new data and randomly initialized. Researchers have also investigated transfer learning without freezing all initial layers. Researchers also investigated transfer learning by unfreezing initial layers after certain epochs.} We perform transfer learning without freezing any layer. As a result, the training becomes similar to traditional training with transferred initialization. The fully connected head of the model is modified according to the class number of the investigated dataset. Fig. \ref{Chart}(b) presents a graphical representation of traditional transfer learning. 

In the proposed learning with background class, the number of outputs from the fully connected layer becomes equal to the number of classes in the dataset plus one. The extra class is reserved for the background class. We generate the background class considering the classification dataset. Fig. \ref{Chart}(c) presents a graphical representation of the proposed training. Fig. \ref{Chart}(d) presents the meaning of several components in Fig. \ref{Chart}(a)-(c) subplots.

\hlr{Researchers and organizations with limited computation power may not have enough computation resources to train a large neural network with a large dataset within a reasonable time. We propose keeping one background class. That increases the number of classes by one. The number of samples also increases but that increment is much lower than the multitask learning. It is also possible to select an optimal number of background samples.}  

\hlr{Researchers prefer to use any publicly available datasets while developing a pre-trained model for publishing a result. Researchers also prefer to use publicly available models for higher transparency. Therefore, we develop the background class from publicly available image datasets and apply publicly available pre-trained models.}

\begin{table*}[t]
\caption{Size of training, validation, and test data.}
\label{Datasize}
\vskip 0.15in
\begin{center}
\begin{small}
\begin{sc}
\begin{tabular}{|l|c|c|c|c|c|c|c|c|r|}
\hline
Dataset  &\multicolumn{3}{c|}{\hlr{Number of Images in Dataset}} & \hlr{Multitask Pair}& \multicolumn{2}{c|}{\hlr{Background Data}} \\ \cline{2-4} \cline{6-7}
 &    train & validation & test & & \hlr{Source} & \hlr{Size} \\ \hline
STL-10  &4,500&500*&8,000&Oxford-102 &\hlr{Mixed**} & 3,001 \\ \hline
Oxford-102 &5,897&655*&818&STL-10 &\hlr{Mixed**} & 3,001\\ \hline

CUB-200-2011 &5,395&599*&5,794&FGVC Aircraft&\hlr{Mixed**} & 3,001\\ \hline

FGVC Aircraft &4,558 &2,109$^{+}$ &3,333 &CUB-200-2011 &\hlr{Mixed**} & 3,001\\ \hline
K-MNIST  &54,000&6,000*&10,000&EMNIST-Balanced  & EMNIST-Balanced & 23,500\\ \hline
EMNIST-Balanced &101,520&11,280*&18,800&K-MNIST & K-MNIST&70,000 \\ \hline
EMNIST-Byclass &628,139&69,793*&116,323&K-MNIST & K-MNIST& 70,000 \\ \hline

\end{tabular}
\\ * We split the training data to create the validation set. 
\\ **The process of making a common mixed source is available in subsection IV B.
\\ $^{+}$ We transfer 12 images from each validation class folder to the test class folder of the same class.
\end{sc}
\end{small}
\end{center}
\vskip -0.1in
\end{table*}

\begin{table*}[t]
\caption{Classification Accuracies on Different Datasets with Different Training Conditions with Traditional Models.}
\label{Tclass}
\vskip 0.15in
\begin{center}
\begin{small}
\begin{sc}
\begin{tabular}{|l|c|c|c|c|c|r|}
\hline
Dataset & Model & \# Epoch & Transfer Learn. & Multitask Learn. & Proposed \\ \hline

STL-10  &WideResNet-101 &20& 98.40$\pm$ 0.19& 98.42$\pm$ 0.14 & \bf{98.58$\pm$ 0.08} \\ \hline
Oxford-102 &WideResNet-101 &20& 98.78$\pm$ 0.21& \bf{99.03$\pm$ 0.19} & 98.93$\pm$ 0.16 \\ \hline
CUB-200-2011 &Inception-v4 &20& 83.26$\pm$ 2.36& \bf{84.93$\pm$ 1.91}& 83.97$\pm$ 1.74 \\ \hline
FGVC Aircraft &Inception-v4 &20& \bf{86.30$\pm$ 1.20}& 85.12$\pm$ 1.60& 85.91$\pm$ 1.20\\ \hline
K-MNIST  &ResNet-18  &15& 98.14$\pm$ 0.21& 98.41$\pm$ 0.15& \bf{98.60$\pm$ 0.12} \\ \hline
EMNIST-Balanced &ResNet-18 &15& 88.91$\pm$ 1.25& 89.24$\pm$ 1.42& \bf{90.04$\pm$ 1.08} \\ \hline
EMNIST-Byclass &ResNet-18 &15& 87.82$\pm$ 0.99& 87.92$\pm$ 0.89& \bf{88.22$\pm$ 0.79} \\ \hline

\end{tabular}
\end{sc}
\end{small}
\end{center}
\vskip -0.1in
\end{table*}

\begin{table*}[t]
\caption{Classification Accuracies on Different Datasets with Different Training Conditions with Transformer (ViT‑L/16).}
\label{TTransformer}
\vskip 0.15in
\begin{center}
\begin{small}
\begin{sc}
\begin{tabular}{|l|c|c|c|c|c|r|}
\hline
Dataset & \# Epoch & ViT‑L/16 & + Spinal FC & + Background & + (Spinal FC \& Background) \\ \hline

STL-10  &2 &99.50$\pm$ 0.09& 99.58$\pm$ 0.06& 99.63$\pm$ 0.06 & \bf{99.71$\pm$ 0.06}* \\ \hline
Oxford-102  &2 &99.51$\pm$ 0.08& 99.41$\pm$ 0.09& \bf{99.75$\pm$ 0.03} & 99.60$\pm$ 0.05 \\ \hline
CIFAR-10  &3 &98.67$\pm$ 0.25& 98.81$\pm$ 0.19& 98.97$\pm$ 0.18 & \bf{99.05$\pm$ 0.14} \\ \hline
CIFAR-100  &3 &92.88$\pm$ 0.73& 93.11$\pm$ 0.55& \hlr{93.16$\pm$ 0.49}& \hlr{\bf{93.31$\pm$ 0.50}} \\ \hline
Caltech-101  &2 &97.02$\pm$ 0.35& 97.31$\pm$ 0.29& \bf{98.02$\pm$ 0.18}*& 97.69$\pm$ 0.21 \\ \hline
CINIC-10  &2 &95.11$\pm$ 0.14& 95.43$\pm$ 0.19& \bf{95.80$\pm$ 0.07}* & \bf{95.80$\pm$ 0.10}* \\ \hline

\end{tabular}
\\ * SOTA Performance on May 2023.
\end{sc}
\end{small}
\end{center}
\vskip -0.1in
\end{table*}

\section{Results}
To compare transfer learning, multitask learning, and the proposed method, we trained models on STL-10, FGVC Aircraft, CUB-200-2011, Kuzushiji-MNIST, EMNIST-(Balanced, Byclass), and Oxford-102 datasets. The proposed method provides higher accuracy compared to transfer learning. The accuracy is slightly higher on average compared to multitask learning. The time required to train models in the proposed method is significantly lower than the multitask learning. We investigate the proposed method with the transformer on STL-10, CIFAR-10, CIFAR-100, Oxford-102, Caltech-101, and CINIC-10 datasets. While applying the vision transformer, we compare transfer learning, the proposed method, and the proposed method with the SpinalNet fully connected layer.

Many of the datasets do not have labeled testing data. In such a situation, we split the training data. We keep 90\% samples as the training data and 10\% samples as the validation data. Also, we consider the validation data of the original dataset as the test data. We apply the stochastic gradient descent (SGD) optimizer with the cross-entropy loss criterion to train our models \hlr{on all datasets.} In the transfer learning method, the number of outputs in the model is equal to the number of classes on the data in which the model is trained.
We share the fully connected head layer in multitask learning and arrange data accordingly. In the multitask learning method, the number of outputs in the model is equal to the summation of the number of classes on datasets trained together. In the proposed method of training with background class, the number of outputs in the model is equal to the number of classes on the \hlr{dataset} on which the model is trained; plus one for the background class. Table \ref{Datasize} presents sizes of datasets, and \hlr{Table \ref{Tclass} presents classification accuracies of applied models on our investigated datasets.}

\subsection{Transfer Learning and Multitask Learning}
\subsubsection{STL-10, and Oxford-102 Datasets}
The STL-10 dataset \cite{coates2011analysis} contains ten classes. As the dataset contains \emph{bird} and \emph{airplane} classes, multitask learning with CUB-200-2011 \cite{wah2011caltech} and FGVC Aircraft \cite{maji2013fine} datasets can potentially degrade the performance. Therefore, we perform multitask learning with Oxford-102 flower dataset \cite{nilsback2008automated}.

The STL-10 dataset contains 96$\times$96 sized images. We observe higher accuracy with enlarged images \cite{kabir2024transfer}. Therefore the images are resized to 256$\times$256 sized images while training \hlr{traditional} CNN-type networks. We \hlr{observe} that the transformer cannot receive images of different sizes. The investigated transformar model accepts only 224$\times$224 sized image. Therefore the images are resized to 224$\times$224 sized images while training the transformer. We also perform random rotation, random crop, and random horizontal flip augmentations on training images. Oxford-102 contains images of varying sizes. We also resized them to the same size. Both the transfer learning training and multitask learning training are performed over 20 epochs with a learning rate of 0.001. We apply the WideResNet-101 pre-trained model for both learning methods. That model is pre-trained on the ImageNet dataset. We download the WideResNet-101 pre-trained model from the \emph{torchvision} package.

\subsubsection{CUB-200-2011 and FGVC Aircraft}
As the CUB-200-2011 dataset \cite{wah2011caltech} contains several images from the ImageNet, we apply the Inception-v4 pre-trained model on CUB-200-2011, and  FGVC Aircraft datasets \cite{maji2013fine}. We download the Inception-v4 pre-trained model from the \emph{timm} package. The CUB-200-2011 dataset contains images of birds. The FGVC Aircraft datasets contain images of aircraft.  They do not have a class in common.

Images of both datasets are resized to 448$\times$448 sized images while training Inception-v4. We also perform random rotation, random crop, random perspective, random vertical flip, and random horizontal flip augmentations on training images. We train models for 20 epochs with a 0.01 learning rate.

\subsubsection{Kuzushiji-MNIST, EMNIST-(Balanced, Byclass)}
Both the KMNIST \cite{clanuwat2018deep} and the EMNIST \cite{cohen2017emnist} datasets contain images of 28$\times$28 sizes.
We apply the ResNet-18 pre-trained model from \emph{torchvision} package for the classification of handwritten letters and digits datasets as they do not require very large models for good performance. KMNIST is quite different from English letters and digits. Therefore, we perform two multitask learning: 1) learning with KMNIST and EMNIST-Balanced pair, and 2) learning with KMNIST and EMNIST-Byclass pair. We resize images to 120$\times$120 size to achieve slightly higher accuracy. We also perform random perspective, random crop, and random rotation augmentations on the training dataset. Both the transfer learning training and multitask learning training on KMNIST/EMNIST datasets consist of two stages: 1) high learning rate training at 0.01 learning rate for 10 epochs and 2) low learning rate training at 0.001 learning rate for 5 epochs.

\subsection{Training with Background Class}
We generate a background class and commonly use it for all the investigated color image classification datasets. This work also classifies several handwritten letter and digit datasets. Handwritten letter and digit datasets contain grey-scale images. We use characters of one language as the background images for classifying characters of another language.

\subsubsection{Background Class for Color Images}
To develop the background class, we investigate background objects of different images. In this study, we investigate STL-10, Oxford-102, CUB-200-2011, and FGVC Aircraft color image datasets. The background of images in these datasets usually contains land, sky, trees, grass, etc. Backgrounds also contain many other objects, but those objects are not frequent. A few images contain those background patterns. We try to develop the background class from publicly available color image datasets. Fig. \ref{Background} presents a few example images in the background class for color images. The Intel Image Classification dataset is publicly available.  Fig. \ref{Background}(a) presents an image of a forest. That image is collected from the Intel Image Classification\footnote{kaggle.com/datasets/puneet6060/intel-image-classification} dataset. Fig. \ref{Background}(b) presents the color-inverted image of Fig. \ref{Background}(a). We also keep inverted images of many backgrounds for robust model training. The presence of an inverted image can potentially train a more robust model that concentrates on exact textures instead of colors. Moreover, the presence of both original and color-inverted images on the background class can potentially make the domain of the background class larger. Fig. \ref{Background}(c)-(d) present glacier and mountain images, collected from the intel image classification dataset.
Fig. \ref{Background}(e) presents monochromatic background images. Monochromatic background images can potentially reduce color bias on classifications. 

Fig. \ref{Background}(f) presents an image of grass from the GrassClover Dataset\footnote{kaggle.com/datasets/usharengaraju/grassclover-dataset}. GrassClover Dataset contains high-resolution images and a few images contain flowers. We generate a few 512$\times$512 sized flower-free grass images from this dataset by resizing or cropping.
Fig. \ref{Background}(g) presents a satellite image from a dataset\footnote{kaggle.com/competitions/understanding\_cloud\_organization}. The image contains a cloud pattern at a location at a time. We extract multiple images from such satellite images by cropping and resizing and saving them as background images.  Fig. \ref{Background}(h) presents images of Galaxies, collected from Galaxy Zoo dataset\footnote{kaggle.com/datasets/jaimetrickz/galaxy-zoo-2-images}.
Fig. \ref{Background}(i) presents water body from another satellite image dataset\footnote{kaggle.com/datasets/franciscoescobar/satellite-images-of-water-bodies}, and Fig. \ref{Background}(j) presents a wood texture from another dataset\footnote{kaggle.com/datasets/edhenrivi/wood-samples}. We took several images from each dataset to construct the background class.

\begin{table*}[t]
\caption{Ablation Study. The first three rows present Data Ablation on the STL-10 dataset. The fourth row presents the model Ablation on the ResNet18 Model. The proposed method consists of a transfer learning and a background class.}
\label{AblationT}
\vskip 0.15in
\begin{center}
\begin{small}
\begin{sc}
\begin{tabular}{|l|c|c|c|c|c|r|}
\hline
Dataset & Model & \# Epoch & Transfer Learn. & Multitask Learn. & Proposed \\ \hline

STL-10 (Red Ablation) &ViT‑L/16 &2& 98.65$\pm$ 0.15& 98.71$\pm$ 0.15 & \bf{98.79$\pm$ 0.14} \\ \hline
STL-10 (Green Ablation) &ViT‑L/16 &2& 98.80$\pm$ 0.09& \bf{98.86$\pm$ 0.09} & 98.84$\pm$ 0.08 \\ \hline
STL-10 (Blue Ablation) &ViT‑L/16 &2& 99.36$\pm$ 0.07& 99.36$\pm$ 0.09 & \bf{99.40$\pm$ 0.08} \\ \hline
K-MNIST &ResNet-18 &15& 98.05$\pm$ 0.31& 98.30$\pm$ 0.20 & \bf{98.54$\pm$ 0.13} \\ 
&(Layer 4 Ablation)&&&& \\ \hline

\end{tabular}
\end{sc}
\end{small}
\end{center}
\vskip -0.1in
\end{table*}

\hlr{In subsection \ref{BG_gen}, we mentioned that the background class images should not contain any target pattern. CIFAR-100 images contain several classes containing trees, furniture, and clouds. The dataset contains willow, pine, palm, oak, and maple trees as the target class. Moreover, the CIFAR-100 dataset contains seas, plains, mountains, forests, and clouds as the target class. Therefore, we do not consider the Intel Image Classification$^2$ dataset to generate background images. Moreover, we cannot consider any dataset containing images of household furniture as the target class to generate background images. Moreover, we cannot consider any dataset that contains images of clouds to generate the background class. We consider GrassClover Dataset$^3$, Galaxy Zoo Dataset$^5$, Satellite Images of Water Bodies Dataset$^6$, and the Wood Texture Samples Dataset$^7$ to generate the background class for the CIFAR-100 dataset. The size of the dataset for the common color background images contains 3001 images. The CIFAR-100 background dataset contains 2528 images. Link to Kaggle notebooks used to extract images are shared to our GitHub repository, sharing scripts.}

\subsubsection{Background Class for Greyscale Images}
We investigate our proposed method on several handwritten digits and characters datasets. These datasets contain grey-scale images. Color images are of different types. Therefore, performing multitask learning or training color and grey images together can potentially degrade performance. Therefore, we observe both English and Japanese characters. KMNIST and EMNIST datasets have no common pattern. Therefore, we use KMNIST data as the background while training EMNIST datasets. Moreover, we use EMNIST data as the background while training KMNIST datasets. The EMNIST-Balanced dataset is much larger than KMNIST. Therefore, we choose 500 samples from each class of the EMNIST-Balanced dataset as images of the background class while training models on KMNIST.

Training parameters with the background class are the same as the training parameters with transfer learning and multitask learning. We keep the same learning rate, epoch number, optimizer, etc. Table \ref{Tclass} presents classification accuracies on different datasets with different training conditions. 

The proposed method provides superior performance on STL-10, KMNIST, and EMNIST datasets. Performance is slightly lower compared to other investigated methods in fine-grained image classification datasets. We also perform deep feature factorization for STL-10 test images. The target class covered 36.8\% of areas on average for the traditional transfer learning method. The majority portion of the image is misclassified in most situations. The white portion of the background area is classified as an airplane, and the green portion of the background is classified as deer in many images. The target class covered 45.9\% of the areas on average in the proposed method. Although the majority portion of the image is still misclassified, a larger area gets the proper classification. Overall accuracy improvement also indicates uncertainty reduction.

\subsubsection{Training with Transformer}
Large language models (LLMs) are transformer-type neural networks. Transformer architectures have received recent SOTA performances with significant improvement over traditional CNNs \cite{ranftl2021vision, fan2021multiscale}.
Therefore, we investigate transformers on several classification datasets and obtain state-of-the-art (SOTA) performance in most situations. As background classes do not contain any target objects of newly investigated datasets, we apply the same background class. The background class we developed for color images in Table \ref{Tclass}. Table \ref{TTransformer} presents classification results while using the ViT‑L/16 transformer. We investigate the ViT‑L/16 model independently, with the SpinalNet fully connected (Spinal FC) layer \cite{kabir2022spinalnet}, and with the proposed background class. 
We have used the same augmentations with the transformer. However, the investigated transformer model takes 224$\times$224 sized images. \hlr{Therefore,} we ensured that the augmented image meets the size criteria of the transformer. 
We receive SOTA performance on STL-10, Caltech-101, and CINIC-10 datasets. We have uploaded scripts on GitHub. We demonstrated SOTA performances on Kaggle servers. We have provided links to Kaggle notebooks on the GitHub repository.

\subsection{Ablation Study}

We have investigated both Data Ablation and Model Ablation. 

\subsubsection{Data Ablation}
\begin{figure}[ht]

\begin{center}
\centerline{\includegraphics[clip, trim=0.3cm 0.3cm 14.5cm 0.2cm, width=\columnwidth,angle=0]{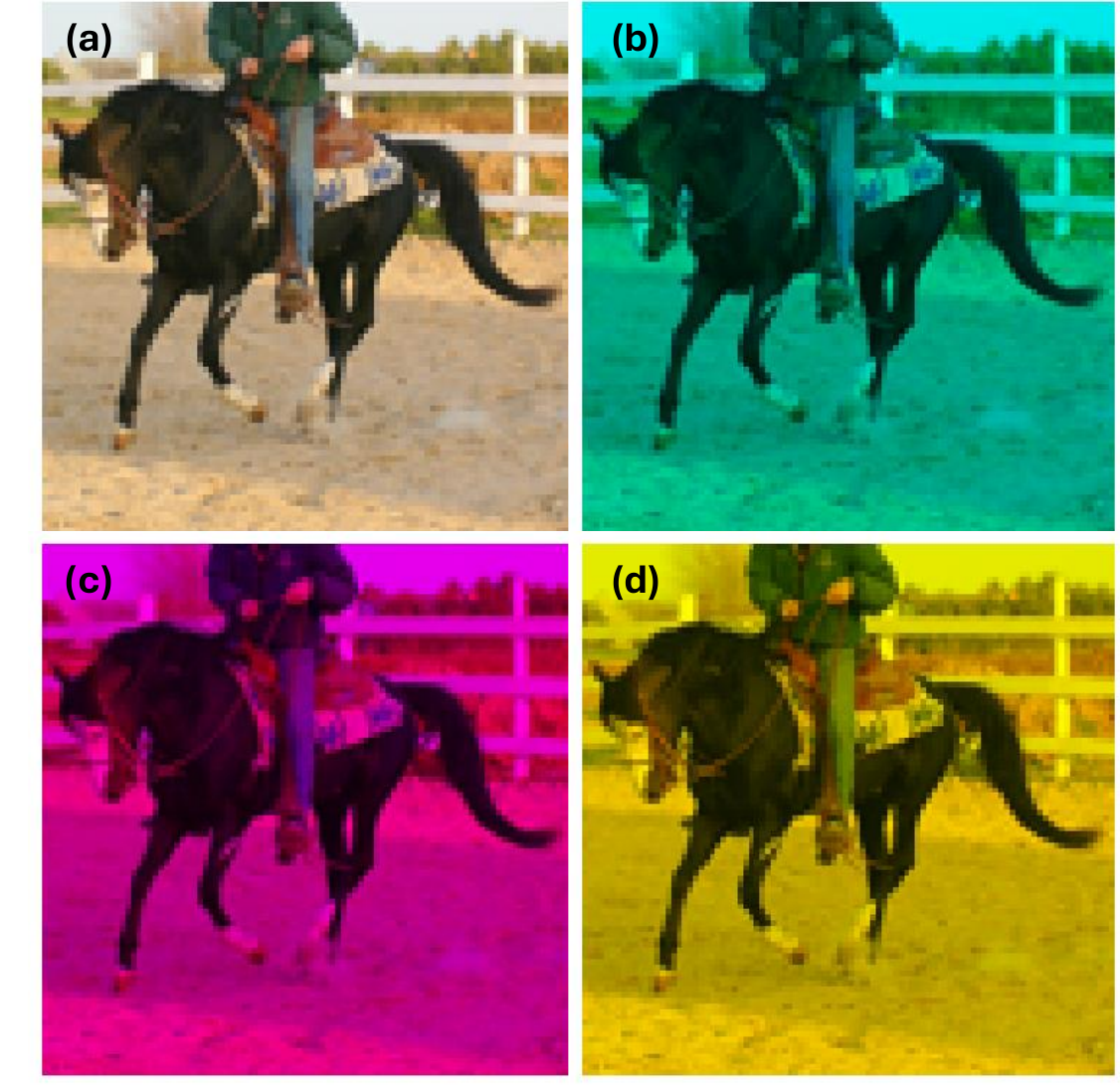}}
\caption{Data Ablation. Original image (a), removal of red component (b), removal of green component (c), and removal of blue component (d).}
\label{DataAblation}
\end{center}
\end{figure}

Fig. \ref{DataAblation} presents the effect of ablation on an image in the STL-10 dataset. Fig. \ref{DataAblation}(a) presents the original image. The image is an image of a horse. Fig. \ref{DataAblation}(b) presents the image after the removal of the red (R) component. Fig. \ref{DataAblation}(c) and Fig. \ref{DataAblation}(c) present images after the removal of the green (G) and blue (B) components respectively. We have applied the ViT‑L/16 transformer model on ablated datasets.

We investigate transfer learning, multitask learning, and training with background classes on the ablated STL-10 dataset. Results are presented in the first three rows of Table \ref{AblationT}. We achieve 99.63\% average accuracy without ablation. Red (R), green (G), and blue (B) ablations provide 98.79\%, 98.74\%, and 99.40\% accuracies on average. According to the result of data ablation, red and green components carry major information. The absence of the red or the green components decreases the accuracy largely, compared to the absence of the blue component. As models are well generalized, the accuracy lowering due to ablation is low.

\subsubsection{Model Ablation}
We investigate layer ablation of the ResNet-18 model on the K-MNIST dataset. The initial layers of ResNet-18 contain four major layers. Each layer contains two basic blocks. Basic blocks contain multiple convolutional layers and batch normalizations. We ablated layer four by forwarding the input of layer four to the input of the fully connected layer. Smaller CNNs, such as VGG-5 can provide near SOTA performance on K-MNIST datasets \cite{kabir2022spinalnet}. Therefore, deletion of the layer four, followed by re-training does not degrade the performance significantly. The average performance is degraded by 0.09\%, 0.11\%, and 0.06\% respectively for transfer learning, multitask learning, and the proposed method. The fourth row of Table \ref{AblationT} presents the accuracies of trained models after the layer ablation.

\subsection{\hlr{Test Model on Noisy Data}}
\hlr{Although we trained models on clean data, the test image may contain noise. The presence of noise often changes the prediction result. Therefore, models need to be robust against common noise patterns. Researchers have recently proposed several datasets containing images with noises. One of the most popular datasets of noisy images is the CIFAR-10C dataset \cite{hendrycks2018benchmarking}. That dataset contains corrupted test images of the CIFAR-10 dataset. The CIFAR-10C dataset contains images with different corruptions. Table \ref{Noisy} presents the accuracies of trained models on the CIFAR-10C dataset. We train the ViT‑L/16 model with and without the proposed data arrangement. Model training with background class brought superior performance for all the combinations. There are nineteen corruptions on the dataset. The last combination on the table is the clean test accuracy. Among these twenty combinations, ViT‑L/16 with background class received the highest accuracy in eight combinations. ViT‑L/16 with background class and SpinalNet fully connected layer received the highest accuracy in twelve combinations. We perform the evaluation on Kaggle servers for transparency. Also, links to Kaggle servers are available in the GitHub repository of the paper.} 

\begin{table}[t]
\caption{\hlr{Test Accuracy of Trained Models on CIFAR-10C Datasets.}}
\label{Noisy}
\vskip 0.15in
\begin{center}
\begin{small}
\begin{sc}
\begin{tabular}{|l|c|c|c|c|c|}
\hline
\rotatebox{90}{\parbox[c]{2.2cm}{ CIFAR-10C Corruption Type}}  & \rotatebox{90}{ViT‑L/16} & \rotatebox{90}{+ Spinal FC } & \rotatebox{90}{+ Background } & \rotatebox{90}{\parbox[c]{2.2cm}{ + (Spinal FC \& Background)}} \\ \hline

brightness         &98.76&98.68&\bf{99.03}&98.96 \\ \hline
contrast           &98.86&98.69&98.82&\bf{98.94} \\ \hline
defocus\_blur      &98.83&98.64&\bf{98.96}&98.83 \\ \hline
elastic\_transform &97.62&97.47&97.85&\bf{97.94} \\ \hline
fog                &98.75&98.69&98.88&\bf{98.93} \\ \hline
frost              &98.26&98.32&\bf{98.71}&\bf{98.71} \\ \hline
gaussian\_blur     &98.83&98.63&\bf{98.92}&98.86 \\ \hline
gaussian\_noise    &97.12&97.43&97.29&\bf{97.74} \\ \hline
glass\_blur        &89.70&90.27&\bf{90.37}&89.41 \\ \hline
impulse\_noise     &98.00&97.52&\bf{98.47}&98.45 \\ \hline
jpeg\_compression  &96.47&96.87&96.34&\bf{96.92} \\ \hline
motion\_blur       &98.05&97.78&98.13&\bf{98.29} \\ \hline
pixelate           &98.30&98.43&98.57&\bf{98.62} \\ \hline
saturate           &97.70&97.62&97.98&\bf{98.18} \\ \hline
shot\_noise        &97.88&97.99&98.00&\bf{98.36} \\ \hline
snow               &97.95&98.15&98.49&\bf{98.61} \\ \hline
spatter            &98.54&98.34&\bf{98.72}&98.68 \\ \hline
speckle\_noise     &98.01&98.08&97.93&\bf{98.28} \\ \hline
zoom\_blur         &97.58&97.31&97.95&\bf{98.08} \\ \hline
CIFAR-10 Test Set  &98.86&98.82&\bf{98.98}&98.92 \\ \hline

\end{tabular}
\end{sc}
\end{small}
\end{center}
\vskip -0.1in
\end{table}

\subsection{Discussion}
According to discussions with colleagues and reviewers' concerns, the current process of background class generation requires manual labor. Finding datasets containing potential background images is not laborious. However, we often need to check whether an image contains the pattern of a target class or not. The requirement for human labor becomes high when the number of background class images is high. Future researchers can potentially extract background images from datasets where all objects in images are previously labeled. For example, the cityscapes dataset \cite{cordts2016cityscapes}. A script can select images based on labels. Background image class of the bicycle classification problem can potentially contain all images from the cityscapes dataset; except images containing bicycles. 

The performance may vary based on the size of the dataset. Training with a larger amount of data brings better generalization \cite{tian2022comprehensive}. In a large dataset, some images can potentially contain different background patterns. As a result, the model may not consider the background pattern as the pattern of the class. Generalization increases dramatically with the slight increase in data size when the dataset is small. However, the improvement of generalization becomes slight due to an increase in data size when the dataset is large \cite{morcos2018importance}. The data size needs to be very large to achieve good generalization. A larger data size increases training time significantly. The proposed background class method requires less data to bring good generalization. In some situations, finding a large dataset is challenging. It is not convenient to increase data size significantly to receive some different backgrounds. Moreover, neural network training considers statistical relationships \hlr{between example inputs and example outputs}. A few images with different backgrounds can potentially make an insignificant difference in training. When the dataset is large, the user of the proposed method can increase the background class size to bring good generalization.

\hlr{We received SOTA performance on the STL-10 dataset using ViT models from Hugging Face. These ViT models are trained on the ImageNet-21k dataset and fine-tuned on the ImageNet-1k dataset. The STL-10 dataset contains labeled examples of ImageNet. Therefore, there is a possibility that some test images of the STL-10 dataset are the training images of the ImageNet dataset. Consequently, we do not claim the SOTA performance on the STL-10 dataset.}

\hlr{Domain adaptation is a closely related research area. Readers can potentially confuse the proposed method with domain adaptation. In the proposed method, we are proposing a background class to improve generalization, and the domain adaptation method makes the neural network robust against any change in the domain. The proposed approach can potentially make the models more robust against any change in domain. However, recently proposed  SoTA performing domain adaptation methods apply quite different neural network training methods with added modules \cite{wang2022continual, chen2022contrastive}.}

\hlr{In this work, we are adding an extra class, known as the background class. The training time becomes longer as the background class contains a large number of images. As mentioned in Table I, the size of training data increases by 30\% to 70\%. However, the computation complexity during the execution is almost the same as the traditional counterpart. If the traditional method classifies images into N number of classes, the proposed method classifies images into N+1 number of classes. Therefore, the number of computations in the initial layers is the same. The number of computations in the fully connected head layers is increased by (100/N) percentage. Moreover, we can discard the incoming weights and biases of the background class and consider the maximum value of traditional N classes. Then the computation becomes the same as the traditional counterpart.}

\section{Potential Future Applications}
Although we apply both traditional CNNs and transformers, we obtain CAM from only CNNs. There exist several recent approaches trying to get smooth CAM from transformers \cite{li2023transcam, qiang2022attcat}. Future researchers can potentially find better CAM on vision transformers and investigate the effect of background class with CAM on transformers.

\hlr{We have applied the background class to a few class classification problems. We demonstrated our method in a few datasets. Many other classification and detection problems face background objects. Background patterns are also prevalent in remote sensing \cite{hong2023cross, hong2024spectralgpt, data2024multimodal}. Researchers may apply the concept of background class on remote sensing in the near future.}

Future researchers may use the proposed background class in other multiclass problems and new datasets \cite{pannattee2024american, maitra2024virtual, lambert2024trustworthy}. Cameras are attached to many industrial equipment. Various computer vision models compute important results from photos taken by those cameras. Such cameras usually capture a limited number of background patterns. Researchers can potentially develop datasets with background patterns for customized detection and classification scenarios and apply the proposed model for superior performance.

Deep learning has become a hot research topic in recent decades \cite{lecun2015deep, mishra2024lsco}. Researchers and top organizations are proposing new state-of-the-art models and approaches almost every year. We have applied the proposed method while training several ResNet-type convolutional neural networks and vision transformers. Researchers in the future can potentially apply the proposed method with a superior future model and get superior performance. Researchers in the future can potentially apply other performance enhancement methods to achieve superior performance \cite{ofori2024defending}. Such methods can potentially be customized data augmentation, novel training methodology, adversarial training, etc.

\hlr{The test image never becomes the same as a training image in practical situations. We apply various image augmentation to training images to improve generalization. We have applied different image flips and rotations to augment data. Such augmentation makes the trained model robust against flips and rotations. However, the noise can change the prediction result \cite{hong2018augmented}. Even one of Google's high-performing models predicted a panda as a gibbon \cite{goodfellow2014explaining}.} Researchers add noise as augmentation and perform adversarial training to overcome such issues \cite{kabir2018partial, cubuk2019autoaugment}. Future researchers can potentially perform robust augmentation or adversarial training with the proposed background class and receive superior performance.

\section{Conclusion}
In this work, we have proposed background classes to reduce class activation uncertainty without significantly increasing the training time. Our theoretical study also indicated an improved generalization while using an optimal background class. We have written a short methodology for developing background classes. According to our experiments on several publicly available datasets, we have statistically received superior performance while using the background class. We have received significant SOTA or near-SOTA performances in several datasets by applying both the ViT‑L/16 transformer and the background class. We have also discussed potential future applications and \hlr{potential} improvements.


\bibliographystyle{IEEEtran}
\bibliography{Ref}

\end{document}